\documentclass[letterpaper, 10 pt, conference]{ieeeconf}   

\IEEEoverridecommandlockouts                              % This command is only needed if 
\usepackage[table,xcdraw]{xcolor}
\usepackage{todonotes}
\usepackage{booktabs}
\usepackage{soul} %for text highlight
\usepackage{svg}
\usepackage{cite}
\usepackage{amsfonts}
\usepackage{multirow}
\usepackage{makecell}

\usepackage{float}
\let\labelindent\relax
\usepackage{enumitem}
\usepackage{mathtools}
\usepackage{epsfig}
\usepackage{epstopdf}
\usepackage{amsmath,hyperref}
\usepackage[linesnumbered, ruled, vlined]{algorithm2e}

\makeatletter
\newcommand*{\rom}[1]{\expandafter\@slowromancap\romannumeral #1@}
\makeatother

\SetAlCapFnt{\footnotesize}
\SetAlCapNameFnt{\footnotesize}
\usepackage{balance}
\usepackage{xcolor}

\SetCommentSty{mycommfont}

\title{\LARGE \bf
Safe Whole-Body Loco-Manipulation via Combined Model and Learning-based Control}
\author{Alexander Schperberg$^{{\dagger}{}*}$, Yeping Wang$^{\ddagger}$, and Stefano Di Cairano$^{\dagger}$%
    \thanks{$^{\dagger}$Alexander Schperberg, and Stefano Di Cairano are with Mitsubishi Electric Research Laboratories (MERL), Cambridge, MA, 02139, USA. {\tt\small \{schperberg, dicairano\}@merl.com}}
    \thanks{$^{\ddagger}$Yeping Wang was with Mitsubishi Electric Research Laboratories (MERL) during part of the work of this paper, Cambridge, MA, 02139, USA. {\tt\small yeping@cs.wisc.edu}}
    \thanks{$^*$Corresponding author}
}

\begin{document}
\maketitle
\thispagestyle{empty}
\pagestyle{empty}

%%%%%%%%%%%%%%%%%%%%%%%%%%%%%%%%%%%%%%%%%%%%%%%%%%%%%%%%%%%%%%%%%%%%%%%%%%%%%%%%
\begin{abstract}
Simultaneous locomotion and manipulation enables robots to interact with their environment beyond the constraints of a fixed base. However, coordinating legged locomotion with arm manipulation, while considering safety and compliance during contact interaction remains challenging. To this end, we propose a whole-body controller that combines a model-based admittance control for the manipulator arm with a Reinforcement Learning (RL) policy for legged locomotion. The admittance controller maps external wrenches—such as those applied by a human during physical interaction—into desired end-effector velocities, allowing for compliant behavior. The velocities are tracked jointly by the arm and leg controllers, enabling a unified 6-DoF force response. The model-based design permits accurate force control and safety guarantees via a Reference Governor (RG), while robustness is further improved by a Kalman filter enhanced with neural networks for reliable base velocity estimation. We validate our approach in both simulation and hardware using the Unitree Go2 quadruped robot with a 6-DoF arm and wrist-mounted 6-DoF Force/Torque sensor. Results demonstrate accurate tracking of interaction-driven velocities, compliant behavior, and safe, reliable performance in dynamic settings.
\end{abstract}

\section{Introduction}
Legged robots with manipulator arms have the potential to not only overcome uneven terrain or cluttered human environments, but can also expand the manipulator overall workspace, enabling greater range of operation and achievable tasks from search and rescue to human-robot collaboration. Modeling such complex systems is inherently difficult, and existing approaches typically either solve a large-scale optimization problem that jointly address locomotion and manipulation (loco-manipulation) \cite{sentis_synthesis_2005}, or rely on learning-based methods \cite{liu2024visual}, which either lack formal safety guarantees or require enormous amounts of data as the number of controllable joints increase. Due to such complexity, even learning-based methods typically opt for a hierarchical architecture \cite{pan2024roboduet}, separating the arm and leg policies, which effectively limits performance as both policies do not coordinate together to achieve the desired task. Although some of the state-of-the-art algorithms are starting to employ a more unified policy, requiring only a single training step \cite{fu2022deep}, controlling contact forces adequately remains an active field of research in the loco-manipulation domain. This is especially true as many approaches focus primarily on tracking end-effector poses. Further, while Reinforcement Learning (RL) has shown success in contact-rich tasks, it typically manages contact forces implicitly, without explicitly regulating them. However, the ability to regulate forces is critical to enable compliant behavior essential for kinesthetic teaching, safe human-robot interaction, avoid damaging the robot or others during contact, and for allowing optimization of body posture to improve the force manipulability ellipsoid, which facilitates the generation of greater task-aligned forces. 

\begin{figure}[t]
    \centering
    \includegraphics[width=0.99\columnwidth]{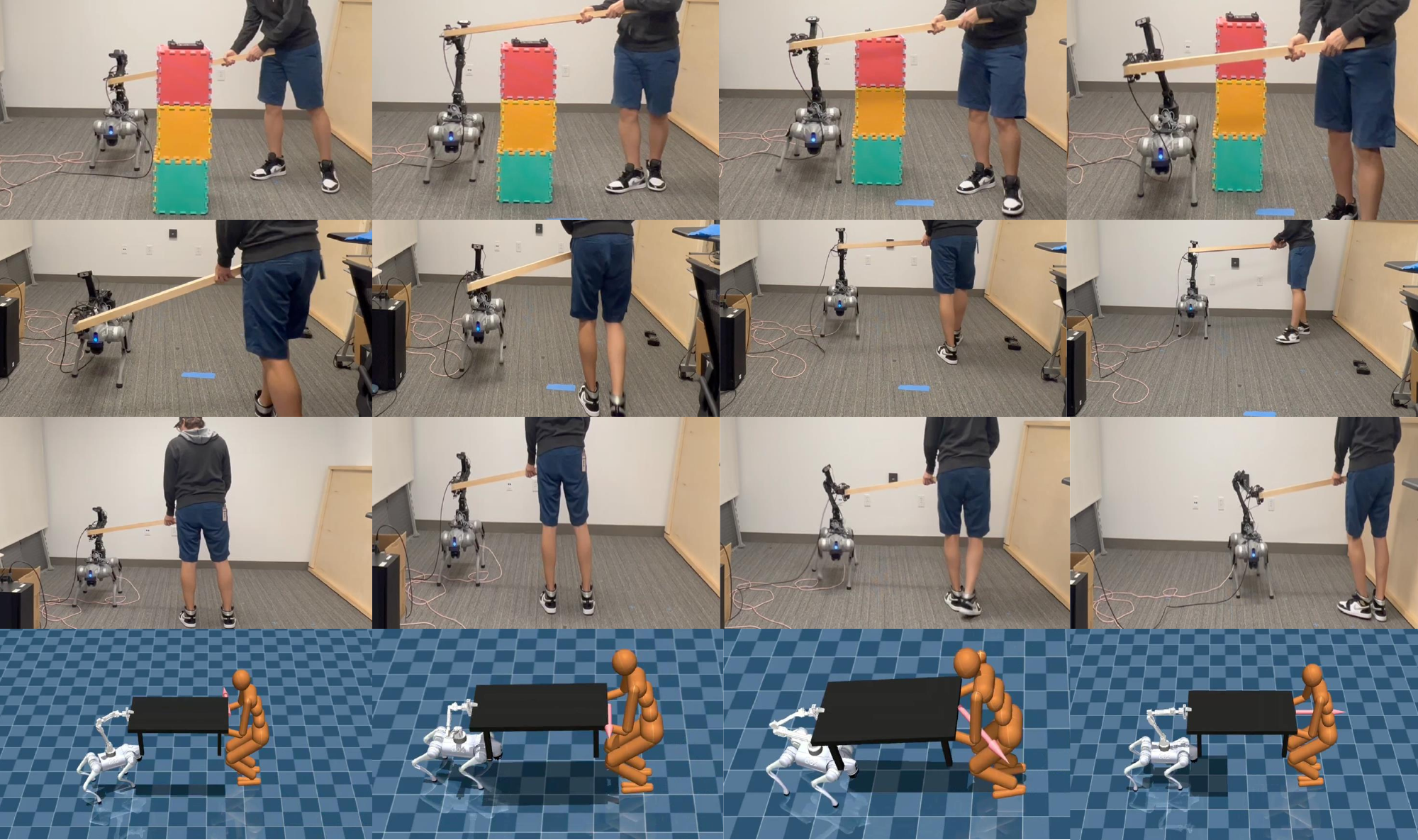}
    \caption{We present a combined model and learning-based whole-body controller that enables safe collaboration between a human and a quadruped robot with manipulator arm in both simulation and hardware. } 
    \label{fig:intro}
\end{figure}

Our work falls within the domain of force control for whole-body loco-manipulation. Although some works account for contact forces \cite{portela2024}, they often do not directly measure them using Force/Torque (F/T) sensors, relying instead on simple spring-damper models for estimation. Especially as simulators struggle to model such complex contact forces, we argue that methods which employ F/T sensors are critical for safety and accuracy. More importantly, as many existing methods overlook safety considerations or lack formal guarantees in their controller design---as some employ pure learning-based policies \cite{portela2024, zhi2025learning}, we aim for a different approach. We combine a model-based admittance controller for the manipulator arm with a learning-based RL policy for legged locomotion. The model-based design for the arm was chosen not only because it directly contacts the human or object, demanding higher safety and accuracy requirements, but also because it enables us to enforce and guarantee safety. In our case, this is achieved through a Reference Governor (RG) \cite{ref_gov1}. While the arm and leg controllers are modular, our approach still addresses the coupled dynamics of loco-manipulation by incorporating the admittance controller during the RL training procedure. Such controller tracks the end-effector forces and torques and maps them to desired linear and angular velocities of the combined arm-base system through the Jacobian. Finally, since the admittance controller and the RG are state feedback methods, state estimation directly impacts controller performance \cite{saber}. Thus, we formulate a Kalman filter enhanced with neural networks to improve linear and angular velocity estimation. We also make use of the state covariance matrix from the Kalman filter to inform the RL observations during the training procedure, reducing the sim-to-real gap (see. Sec. \ref{sec:RL}).

Overall, we provide the following \textbf{contributions}:
\begin{enumerate}
    \item We present a combined model and learning-based whole-body controller for legged loco-manipulation that considers both the legs and arms for tracking forces and torques explicitly through F/T sensors, while guaranteeing safety.
    \item We enforce safety guarantees in controller design for the arm through a Reference Governor.
    \item We formulate a Kalman filter with neural networks to model complex system dynamics to facilitate linear and angular velocity estimation of the base, which directly improves the stability and accuracy of the combined controller. 
    \item We validate the proposed methods in both simulation and hardware experiments.
\end{enumerate}
\section{Related Work}
The loco-manipulation domain can be classified into the following categories based on their shared approaches: Sec. \ref{no_wbc}: Modifying only the planning and control architecture on the manipulation side while using a standard MPC or RL-based locomotion controller (i.e., no whole-body control); Sec. \ref{model_wbc}: Integrating a model-based whole-body controller that considers the coupling between locomotion and manipulation; Sec. \ref{rl_wbc}: Applying a learning-based approach to whole-body control, which may involve a two-step process where a low-level policy is trained first, followed by a high-level policy using perception and control, or using a unified policy trained in a single step; and Sec. \ref{hybrid_wbc}: Using a approach that combines model-based control for manipulation with a learning-based controller for locomotion, or the reverse. Finally, Sec. \ref{force_wbc} emphasizes force-based controllers for loco-manipulation to enable compliant behavior.
\subsubsection{Control without whole-body consideration}
\label{no_wbc}
Mobile manipulation using default, model-based controllers, typically accessed through the robot's API, is demonstrated in \cite{gamma,yokoyama2024,acari2023}. For example, Zhang \textit{et al.} \cite{gamma} focus on online grasp pose fusion to enable object grasping while the robot is in motion. Yokoyama \textit{et al.} \cite{yokoyama2024} use Bayesian basis functions to collect data for a range of manipulation tasks, including door opening, with locomotion handled by an existing controller. Arcari \textit{et al.} \cite{acari2023} address long-horizon pick-and-place tasks, coordinating high-level commands through the SPOT API to guide the robot in navigating to the object, picking it up, and transporting it to a target location. The limitation of these works is that they decouple the manipulation and locomotion modules, and do not account for the complex dynamics from simultaneous leg and arm motion, reducing performance or time to complete the task.
\subsubsection{Model-based whole-body control}
\label{model_wbc}
The coupling of arms and legs through a pure model-based approach is demonstrated in \cite{sentis_synthesis_2005, posa2014, murphey2012, murooka2015}. In \cite{sentis_synthesis_2005, posa2014, murphey2012}, a whole-body inverse dynamics model is used to enforce stability around a reference trajectory computed using Trajectory Optimization (TO) and tracked by a Model Predictive Controller (MPC). Murooka \textit{et al.} \cite{murooka2015} used contact-based point planning and control for manipulating objects, while accounting for full-body humanoid dynamics. These works tend to use simple robot base/arm dynamics to reduce computation cost for online model-based control which may result in limited performance.
\subsubsection{RL-based whole-body control}
\label{rl_wbc}
End-to-end learning-based whole-body controllers for loco-manipulation has also shown successful results \cite{ liu2024visual, pan2024roboduet, fu2022deep}. Liu \textit{et al.} \cite{liu2024visual} present a Visual Whole-Body Controller (VBC). The VBC employs a low-level goal-reaching policy, which is then leveraged in a second training stage, where a high-level task-planning policy is learned. RoboDuet \cite{pan2024roboduet} is another work that trains two policies: one for locomotion and another for manipulation. They show successful tracking of 6 DoF end-effector pose and velocities for the robot's base. Meanwhile, Fu \textit{et al.} \cite{fu2022deep} avoid the complexities of training two different policies and subsequent reward tuning by proposing a unified policy. The joints of a quadruped robot and manipulator arm are trained simultaneously to track desired end-effector poses and velocities, using domain adaptation techniques. While learning-based approaches are attractive for their generizability and ability to model complex dynamics, their output is not predictable, which makes safety quantification challenging. 
\subsubsection{Combined model and learning-based whole-body control}
\label{hybrid_wbc}
To address the unpredictability of learning-based methods, there are also works that employ a combination of model and learning-based approaches \cite{bellicoso2019,zimmerman2021}.
For example, ALMA \cite{bellicoso2019} is an online motion planning framework that combines a model-based MPC controller for the manipulator arm, which tracks the desired end-effector pose, with a learning-based locomotion policy for maintaining balance. However, the leg and arm controllers operate independently, and the framework does not show dynamically coordinated motions. Similarly, Zimmermann \textit{et al.} \cite{zimmerman2021} also combines model and learning-based components but is limited in effectiveness, as this method lacks tight coordination between the arm and legs. Overall, this modular separation can lead to error propagation between subsystems, resulting in non-smooth and potentially unsafe overall behavior. 
\subsubsection{Implicit force tracking in whole-body control}
\label{force_wbc}
While most works in loco-manipulation focus on tracking reference end-effector poses and base velocities, they do not consider the interplay of contact forces to allow for compliant behavior. Such compliance is beneficial for kinesthetic teaching, preventing damage to the robot or its environment, and facilitating human-robot collaboration. Only a few works have explored this domain \cite{portela2024, Ji_2023, ma_2023, haarnoja2024, zhi2025learning, schperberg2023adaptive}. For example, Portela \textit{et al.} \cite{portela2024} train an RL policy that takes direct target force as input, allowing the teleoperator to modulate the compliance of the manipulator arm. This approach uses an impedance controller, which requires the robot hardware to support direct torque commands without Force/Torque (F/T) sensors. There are also works that focus on learning downstream tasks through force interactions, such as fall recovery using the mounted arm \cite{Ji_2023}, or ball dribbling using the robot's feet \cite{ma_2023, haarnoja2024}. While these works typically have separate policies for the arm and legs to enable force-based loco-manipulation control, Zhi \textit{et al.} \cite{zhi2025learning} demonstrate a unified policy that integrates force and position control without F/T sensor requirements. The policy employs a force estimator to predict external forces based on the offsets between target and current positions (\textit{i.e.}, spring-based models). Although these works adopt a force-centric approach to loco-manipulation, their ability to track forces remains limited, largely due to the absence of F/T sensors, which makes them inadequate for tasks requiring precise and accurate force control. Importantly, they do not provide safety guarantees in their control design.

\section{Methods}

\begin{figure}[t]
    \centering
    \includegraphics[width=0.99\columnwidth]{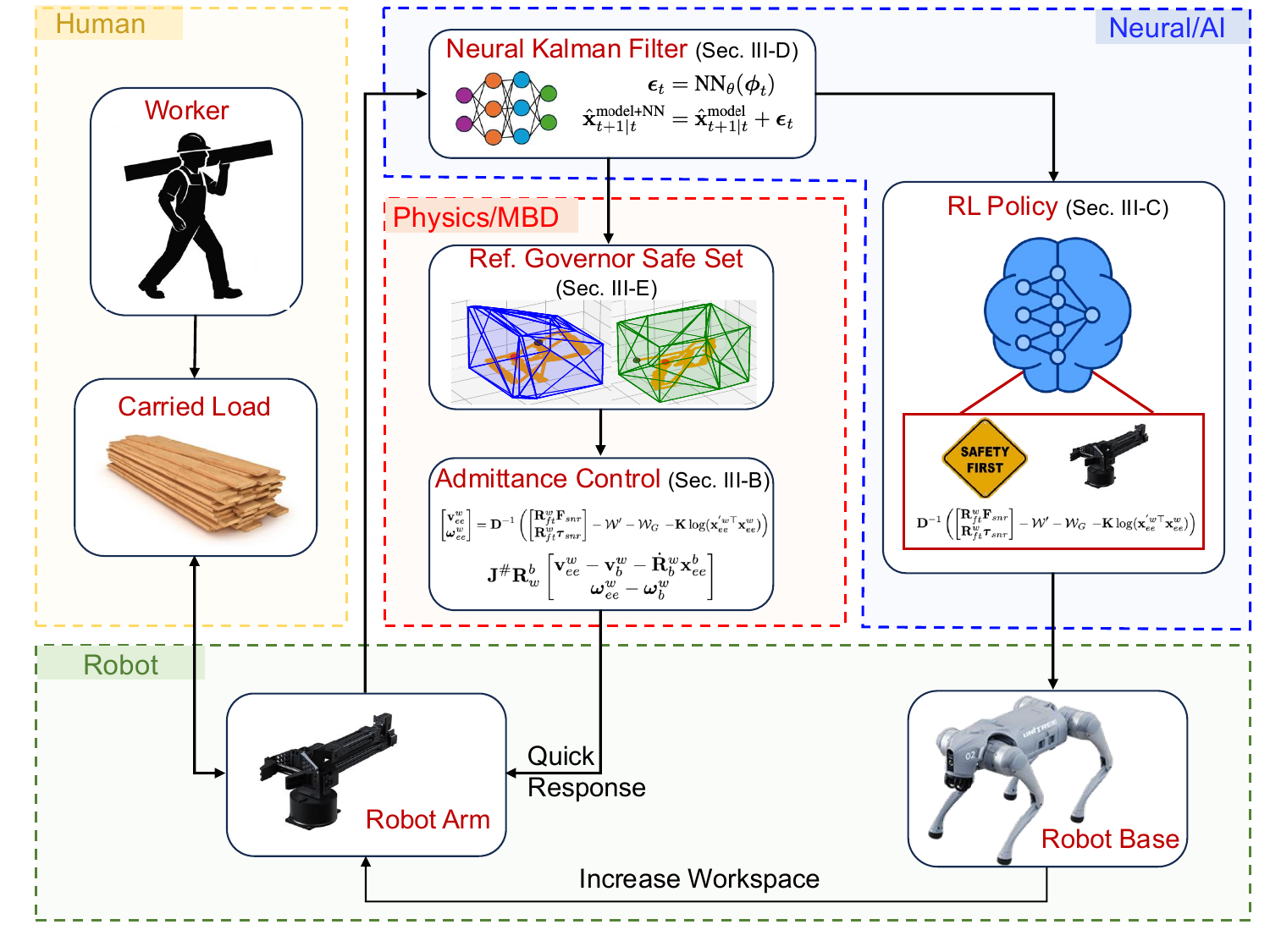}
    \caption{Our framework architecture that combines neural/AI methods with Model Based Designs (MBD) for safe human-robot collaboration.} 
    \label{fig:overview}
\end{figure}
%To ground the problem, consider a human co-transporting a load with a quadruped robot equipped with a robotic arm. The arm grasps one side of the object and maintains stiffness along the vertical $z$ axis to provide stability. At the same time, the robot must respond \emph{compliantly} to forces applied at the end-effector in the horizontal $x$-$y$ plane, so that the object can move smoothly in the direction guided by the human. 
To ground the problem, consider the example application of a human co-transporting a load with a quadruped robot equipped with a robotic arm. The arm grasps the object and must react \emph{compliantly} to wrenches applied at the end-effector so it moves along the human's directed motion.
Such motion is then used to steer the robot base, enabling the robot to follow the human with whom the load is carried. As shown in Fig \ref{fig:overview}, we propose a combined model and learning-based whole-body controller to enable such safe human-robot collaboration.

\subsection{Problem Formulation}
We model the arm as a fast, low-inertia, short-range actuator–sensor that performs rapid local corrections, whereas the robot base functions as a slower, high-inertia, long-range body that repositions to sustain the motion and increase the achievable workspace. Operationally, the arm executes the immediate corrective response, and the base subsequently aligns the whole body to preserve smooth, continuous progress.

%Consider a quadruped robot equipped with a robotic arm. The objective is to control both the legs and the arm such that they respond compliantly to external forces applied on the end-effector. 

Specifically, the end-effector should follow a desired linear velocity $\mathbf{v}_{ee}^w \in \mathbb{R}^3$ and angular velocity $\boldsymbol{\omega}_{ee}^w \in \mathbb{R}^3$ in response to an external wrench  $\mathcal{W} \in \mathbb{R}^6$, while also maintaining a reference pose $\mathbf{x}_{ee}^{'w} \in SE(3)$ and a reference wrench $\mathcal{W}' \in \mathbb{R}^6$ obtained from the Reference Governor in Sec. \ref{sec:ref_gov}. We model this behavior in the world frame as:
\begin{align}
\label{eq:admittance}
\mathcal{W} - \mathcal{W}^{'} = \mathbf{K} \log( \mathbf{x}_{ee}^{'w\top} \mathbf{x}_{ee}^w) + \mathbf{D} \begin{bmatrix}
\mathbf{v}_{ee}^w \\
\boldsymbol{\omega}_{ee}^w
\end{bmatrix} 
%+ M \mathbf{\ddot{x}}_{ee}^w
\end{align}
where $\mathbf{K}$ and $\mathbf{D}$ correspond to the stiffness and damping coefficients which are diagonal with dimensions $\mathbb{R}^{6 \times 6}$ and $\mathbf{x}_{ee}^w \in SE(3)$ is the current pose of the end-effector. The term $\log( \mathbf{x}_{ee}^{'w \top} \mathbf{x}_{ee}^w) \in \mathbb{R}^6$ represents the linear and angular difference between the desired $\mathbf{x}_{ee}^{'w}$ and current $\mathbf{x}_{ee}^{w}$ end-effector poses. 

In an example where a human and a robot are collaboratively moving a table (bottom of Fig. \ref{fig:intro}), assuming the $z$-axis is pointing upward, the end-effector should be compliant in the $xy$-plane and more stiff in the $z$ direction and in rotation.

\subsection{Whole-Body Admittance Control} \label{sec:arm_controller}
 We consider a F/T sensor mounted at the wrist, between the end-effector and the arm, see Fig. \ref{fig:hardware_validation}. The sensor data $[\mathbf{F}_{snr} \in \mathbb{R}^3, \boldsymbol{\tau}_{snr} \in \mathbb{R}^3] $ is expressed in the F/T sensor's frame. Using Eq. \ref{eq:admittance}, we receive the target linear and angular velocity of the end-effector:
\begin{multline} \label{eq:compute_ee_vel_world}
    \begin{bmatrix}
\mathbf{v}_{ee}^w \\
\boldsymbol{\omega}_{ee}^w
\end{bmatrix}  = \mathbf{D}^{-1} \left(
\begin{bmatrix} 
\mathbf{R}_{ft}^w \mathbf{F}_{snr} \\
\mathbf{R}_{ft}^w \boldsymbol{\tau}_{snr}
\end{bmatrix} - \mathcal{W}' - \mathcal{W}_G \right.\\
\left. - \mathbf{K} \log( \mathbf{x}_{ee}^{'w\top} \mathbf{x}_{ee}^w) 
\right)
\end{multline}
where $\mathcal{W}_G \in \mathbb{R}^6$ is the wrench caused by the gravity of the end-effector and $\mathbf{R}_{ft}^w \in \mathbb{R}^{3\times 3}$  is the rotation matrix from the world frame to the F/T sensor frame, defining $\mathcal{W}$ from Eq. \eqref{eq:admittance}.

The end-effector velocity $[\mathbf{v}^w_{ee}, \boldsymbol{\omega}_{ee}^w]$ in the world frame results from the combination of the base velocity $[\mathbf{v}^w_b, \boldsymbol{\omega}_b^w]$ and the relative velocity of the end-effector with respect to the base $[\mathbf{v}^b_{ee}, \boldsymbol{\omega}_{ee}^b]$:
\begin{align} \label{eq:velocites}
\begin{split}
\mathbf{v}^w_{ee} &=  \mathbf{v}^w_b + \dot{\mathbf{R}}^w_b \mathbf{x}_{ee}^b + \mathbf{R}^w_b \mathbf{v}_{ee}^b \\
\boldsymbol{\omega}_{ee}^w &= \boldsymbol{\omega}_{b}^w + \mathbf{R}_b^w \boldsymbol{\omega}_{ee}^b
\end{split}
\end{align}
where $\mathbf{R}_b^w$ is the rotation matrix from the world frame to the robot base's frame. We note that the Eq. \ref{eq:velocites} is derived from taking the derivative on the end-effector position and orientation from the world to end-effector frame, which includes the contribution of the velocities from both the arm itself and the base. Afterwards, we map the desired velocity from the base to end-effector frame $[\mathbf{v}^b_{ee}, \boldsymbol{\omega}_{ee}^b]$ to the joint space:
\begin{align} \label{eq:jacobian}
\mathbf{\dot{q}}_{arm} = \mathbf{J}^{\#} \begin{bmatrix}
\mathbf{v}^b_{ee} \\ 
\boldsymbol{\omega}_{ee}^b
\end{bmatrix}
\end{align}
where $\mathbf{J}^{\#} = \mathbf{J}^\top \left( \mathbf{J} \mathbf{J}^\top + \lambda^2 \mathbf{I} \right)^{-1}$ is the damped pseudo-inverse of the arm's Jacobian matrix $\mathbf{J}$ and $\lambda$ is a damping factor. Substituting Eq. \ref{eq:velocites} into \ref{eq:jacobian}, the desired arm joint velocities are:
\begin{align}
    \dot{\mathbf{q}}_{arm} &= \mathbf{J}^{\#}	
 \mathbf{R}_w^{b} 
                \begin{bmatrix}
                \mathbf{v}_{ee}^w - \mathbf{v}^w_b - \dot{\mathbf{R}}^w_b \mathbf{x}_{ee}^b \\
                \boldsymbol{\omega}_{ee}^w - \boldsymbol{\omega}^w_b
                \end{bmatrix}	
\end{align}
where $[\mathbf{v}^w_{ee}, \boldsymbol{\omega}_{ee}^w]$ is computed using Eq. \ref{eq:compute_ee_vel_world}, $[\mathbf{v}^w_b, \boldsymbol{\omega}_b^w]$ is estimated from the Kalman filter described in Sec. \ref{sec:state_estimation}, $[\mathbf{R}_b^w, \mathbf{\dot{R}}_b^w]$ is measured by the gyroscope mounted at the robot's base, and $\mathbf{x}_{ee}^b$ is computed using the arm's forward kinematics. 

\subsection{Reinforcement Learning Policy for Base Motion} 
\label{sec:RL}
We train a RL policy to control the robot's base in coordination with the arm's admittance controller described in Sec.~\ref{sec:arm_controller}. Building on the MuJoCo Playground PPO setup \cite{zakka2025mujoco}, originally developed for pure locomotion using the Go1 legged robot, we adapt it for the Go2 robot with a manipulator arm and revise the objective to track end-effector arm velocity instead of base velocity. Hyper-parameters, underlying networks, and number of training steps were kept similar to \cite{zakka2025mujoco}. An example of evaluating our policy in simulation is shown at the bottom of Fig. \ref{fig:intro}, carrying a table in coordination with a human. The observation space includes: 1) gravity projected into the body frame, 2) leg joint positions and velocities, 3) linear and angular velocity of the base, 4) the previous action, and 5) the desired end-effector velocity $[\mathbf{v}^w_{ee}, \boldsymbol{\omega}_{ee}^w]$ received from our model-based admittance controller, which runs at each time step of the RL training procedure. Following MuJoCo Playground \cite{zakka2025mujoco}, the action space consists of leg joint position commands $\mathbf{q}_{\text{leg}}\in \mathbb{R}^{12}$ which are scaled and added to the default joint positions $\mathbf{q}_{\text{default}}\in \mathbb{R}^{12}$:
\begin{equation}
    \mathbf{q}_{\text{leg}} = \mathbf{q}_{\text{default}} + k_a \boldsymbol{a}
\end{equation}
where $k_a$ is the action scale on the raw output of the network $\boldsymbol{\alpha}$.  
The reward is a weighted sum of multiple terms, detailed in Table \ref{table:reward-functions}. The main rewards encourage the robot to follow the desired end-effector velocity. 
In addition to motion tracking, the reward includes terms that promote energy efficiency, such as penalties on joint torques and joint velocities. 
Finally, several shaping rewards and penalties are used to guide specific aspects of locomotion behavior, such as encouraging proper foot swing and reducing foot slip. These components help the policy produce goal-directed, efficient, and physically consistent motion. It is important to note that, during both training and evaluation, we do not provide the policy with the ground-truth base velocities from MuJoCo. Instead, we model realistic state estimation uncertainty by replacing the perfect simulator values with noisy estimates derived from our state estimation procedure in Sec.~\ref{sec:state_estimation}, to help match the state estimates likely received from the real robot. Specifically, at each time step $t$ we take the raw base velocity values from MuJoCo and add zero-mean Gaussian noise $\boldsymbol{\eta}_t \sim \mathcal{N}(\mathbf{0}, \mathbf{P}_t)$, where $\mathbf{P}_t$ is the state covariance matrix received from the update step from our state estimator. This sampling procedure preserves both the variance and cross-correlations of the estimated states, ensuring that the learned policy experiences realistic velocity estimation noise rather than relying on perfect simulator knowledge. 

\begin{table}[h!]
\centering
\caption{Reward Functions}
\label{table:reward-functions}
\begin{tabular}{p{1.4in}p{1.75in}}
\hline \rule{0pt}{1.05\normalbaselineskip}%
\textbf{Reward}                  & \textbf{Expression}            
\\ \hline \rule{0pt}{1.05\normalbaselineskip}%
Velocity Tracking
\\ \hline \rule{0pt}{1.05\normalbaselineskip}%
End-Effector Lin Vel Tracking         & $ \phi(\mathbf{v}_{ee}^{'w} - \mathbf{v}_{ee}^{w})$                        \\               
End-Effector Ang Vel Tracking        & $ \phi(\boldsymbol{\omega}_{ee}^{'w} - \boldsymbol{\omega}_{ee}^{w})$     %H\\
%Base Lin Vel Tracking         & $ %\phi(\mathbf{v}_{b, xy}^{'w} - \mathbf{v}_{b, %xy}^{w})$                       
%\\ Base Ang Vel Tracking        & $ \phi(\boldsymbol{\omega}_{ee, z}^{w*} - \boldsymbol{\omega}_{ee, z}^{w})$        
\\ \hline \rule{0pt}{1.05\normalbaselineskip}%
Energy Saving
\\ \hline \rule{0pt}{1.05\normalbaselineskip}%
Joint Torque (Penalty)           & $\|\boldsymbol{u}\| + \left|\boldsymbol{u}\right|$ \\
Action Rate  (Penalty)           & $\|\mathbf{a}_t - \mathbf{a}_{t-1}\|^2$      \\ 
Energy Consumption  (Penalty)    & $ \left| \dot{\boldsymbol{q}}_\text{leg} \right|^\top \left| \boldsymbol{u} \right| $    
\\ \hline \rule{0pt}{1.05\normalbaselineskip}%
Behavior
\\ \hline \rule{0pt}{1.05\normalbaselineskip}%
Pose Deviation (Penalty)                  & $\phi(\boldsymbol{q}_\text{leg} - \boldsymbol{q}_\text{default})$                            \\
Stand Still (Penalty)            & $ \left|\boldsymbol{q}_\text{leg} - \boldsymbol{q}_\text{default} \right| \cdot  \mathbf{1}_{\|\mathbf{v}_{ee}^w \| {<} 0.01} $    \\
Lin. Velocity in Z (Penalty)   & $ v_{b,z}^2$                                                      \\[1mm] 
Ang. Velocity in XY (Penalty) & $ \omega^2_{b,x} + \omega^2_{b,y}$ \\[1mm]
Feet Airtime                     & $ (\mathbf{t}_\text{air} - 0.1)^\top \mathbf{c}^\text{first} \cdot \mathbf{1}_{\|\mathbf{v}_{ee}^w \| {>} 0.01} $ \\[1mm] 
Swing Clearance (Penalty)                   & $ \sum_i \left| h_i - h_{\text{max}} \right| \cdot \left\| \mathbf{v}_{i,xy} \right\|
$          \\[1mm] 
Swing Height (Penalty)                      & $  \sum_i \left( \frac{h_i^\text{peak}}{h_{\text{max}}} - 1 \right)^2 \cdot c_i^\text{first} \cdot \mathbf{1}_{\|\mathbf{v}^w_{ee}\| > 0.01}$    \\[1mm] 
\rule{0pt}{1.05\normalbaselineskip}%
Feet Slip   (Penalty)                & $ \sum_i c_i \|\mathbf{v}_{i,xy}\|^2\cdot \mathbf{1}_{\|\mathbf{v}_{ee}^w \| {>} 0.01}$                                          \\ 
Termination (Penalty)            & $1_{\mathbf{z}_b \cdot \mathbf{z}_w < 0}$   \\[1mm] 
\hline

\multicolumn{2}{p{3.3in}}{
\rule{0pt}{1.05\normalbaselineskip}%
    where $\phi(x) = \exp(-{\|x\|^2}/{\sigma})$, and $\sigma$ is a tunable parameter which controls how sensitive the reward is to tracking error.
    $\boldsymbol{q}_\text{leg}, \boldsymbol{\dot{q}}_\text{leg}, \boldsymbol{u}\in\mathbb{R}^{12}$ denotes the joint position, velocity, and torque at the leg joints, respectively.
    $v_{b,z}, \omega_{b,x}, \omega_{b,y}$ denotes the base's $z$-axis linear velocity, $x$-axis angular velocity, and $y$-axis angular velocity, respectively.
    $\mathbf{v}_{i,xy}$ denotes the velocity of the $i$-th foot in the ground $xy$ plane. $\mathbf{a}_t \in \mathbb{R}^{12}$ denotes the action vector output by the RL policy at time step $t$, corresponding to scaled leg joint position commands or $\mathbf{a}_{t}=k_a\boldsymbol{\alpha}_t$.
    $\mathbf{z}_b$ and $\mathbf{z}_w$ denotes the $z$-axis of the base and world frame, respectively. $\mathbf{t}_\text{air} \in \mathbb{R}^4$ contains the air-phase durations for all four feet. 
    $c_i\in \{0,1\}$ indicates whether the $i$-th foot is in contact with the ground.
    $c^{\text{first}}_i \in \{0,1\}$ indicates whether the $i$-th foot was previously in the air and makes ground contact in the current frame, \textit{i.e.,} $\mathbf{c}^{\text{first}}$ activates terms only at foot touchdown. $h_i$ denotes the distance of the $i$-th foot from the ground, while $h_\text{max}$ is a hyperparameter specifying the maximum distance and $h_i^{\text{peak}}$ denotes the maximum height reached by the $i$-th foot during its current swing phase. To clarify  notation, in the context of RL training, $\mathbf{v}_{ee}^{'w}$, $\boldsymbol{\omega}_{ee}^{'w}$, represent the desired end-effector velocities, which are the output from the admittance controller, or Eq. \ref{eq:compute_ee_vel_world}, while $\mathbf{v}_{ee}^{w}$, $\boldsymbol{\omega}_{ee}^{w}$ are the current velocities.
} \\ 
\end{tabular}

\end{table}

\subsection{Velocity Estimation and Uncertainty} \label{sec:state_estimation}
Legged robot locomotion produces undesired oscillatory noise, which particularly affects the estimation of the robot's base linear and angular velocities, represented as 
$\hat{\mathbf{x}} = [\mathbf{v}_{b}^{w}, \boldsymbol{\omega}_{b}^{w}]^\top \in \mathbb{R}^6$ (not to be confused with the end-effector position notation $\mathbf{x}_{ee}^w$). As these terms are required to derive the linear and angular velocity of the end-effector $[\mathbf{v}_{ee}^{w}, \boldsymbol{
\omega}_{ee}^{w}]$, error in base velocity estimation will directly affect performance of the admittance controller. Due to the complexity of our system, we leverage a neural network to learn its dynamics using motion capture data as ground truth, and incorporate the resulting model into the Kalman filter to improve state estimation. Specifically, the neural network, $\text{NN}_{\theta}(\phi_{t})$, is parameterized to estimate the difference between our dynamic model and ground truth at time step $t$, where the error is defined by $\boldsymbol{\epsilon}_{t}$. In this regard, $\boldsymbol{\epsilon}_{t}$ has a physical mathematical meaning, i.e., $\boldsymbol{\epsilon}_t\boldsymbol{\epsilon}_t^{\top}=\mathbf{Q}_t$, the covariance of the process noise. Thus, we write the Kalman filter prediction step:
\begin{align}
\hat{\mathbf{x}}_{t+1}^{\text{model}} &= \text{Dyn}(\hat{\mathbf{x}}_t, \mathbf{f}_t) + \mathbf{w}_t \\
\mathbf{y}_t &= h(\hat{\mathbf{x}}_t) + \mathbf{v}_t \\
\boldsymbol{\epsilon}_t &= \text{NN}_{\theta}(\boldsymbol{\phi}_t) \\
\hat{\mathbf{x}}^{\text{model+NN}}_{t+1|t} &= \hat{\mathbf{x}}^{\text{model}}_{t+1|t} + \boldsymbol{\epsilon}_t \\
\mathbf{Q}_t &= \boldsymbol{\epsilon}_t \boldsymbol{\epsilon}_t^\top \\
\mathbf{P}_{t+1|t} &= \mathbf{A}_d \mathbf{P}_{t|t} \mathbf{A}_d^\top + \mathbf{Q}_t
\end{align}
and the update step:
\begin{align}
\mathbf{S}_t &= \mathbf{C}_t \mathbf{P}_{t+1|t} \mathbf{C}_t^\top + \mathbf{R}_t \\
\mathbf{K}_t &= \mathbf{P}_{t+1|t} \mathbf{C}_t^\top \mathbf{S}_t^{-1} \\
\tilde{\mathbf{y}}_t &= \mathbf{y}_t - \mathbf{C}_t \hat{\mathbf{x}}^{\text{model+NN}}_{t+1|t} \\
\hat{\mathbf{x}}_{t+1|t+1} &= \hat{\mathbf{x}}_{t+1|t} + \mathbf{K}_t \tilde{\mathbf{y}}_t \\
\mathbf{P}_{t+1|t+1} &= \mathbf{P}_{t+1|t} - \mathbf{K}_t \mathbf{S}_t \mathbf{K}_t^\top
\end{align}
where $\hat{\mathbf{x}}^{\text{ model}}_{t+1|t}$ is the state from the system model $\text{Dyn}(\cdot)$; $\hat{\mathbf{x}}_{t+1|t}^{\text{model+NN}}$ is the state after model correction from the neural network $\text{NN}_{\theta}(\phi_{t})$; $\mathbf{C}_{t}$ is the observation matrix; $\mathbf{S}_{t}$ is the innovation covariance; $\mathbf{K}_{t}$ is the Kalman gain; $\mathbf{P}_{t}$ is the state estimate covariance; and $\mathbf{y}_{t}$ is the measurement update. $\mathbf{A}_d$ is the discrete-time linearization that is derived from our system dynamics $\text{Dyn}(\cdot)$:
\begin{equation}
\text{Dyn}(\mathbf{x}_k, \mathbf{f}_k) 
=
\begin{bmatrix}
\boldsymbol{\omega}_k + \Delta T \left( \sum_{i=1}^{n} \widehat{\mathbf{I}}^{-1} \left( \mathbf{p}^{i}_{k} \times \mathbf{f}^i_k \right) \right) \\
\mathbf{v}_k + \Delta T \left( \sum_{i=1}^{n} \frac{1}{m} \mathbf{f}^i_k + \mathbf{g} \right)
\end{bmatrix}
\label{eq:reduced_model}
\end{equation}
where $\mathbf{f}_{k}^{i} \in \mathbb{R}^3$, $\mathbf{p}_{k}^{i} \in \mathbb{R}^3$ is the ground reaction force and footstep position in world frame for leg $i$, $\hat{\mathbf{I}} \in \mathbb{R}^{3\times3}$ is the inertia tensor skew-symmetric matrix, $\Delta T$ is the discretized time step, and $\mathbf{g}\in \mathbb{R}^3$ is a constant gravity vector. 

For the measurement model $h(\hat{\mathbf{x}}_{t})$, we use the robot base angular velocity derived from IMU measurements.  We also estimate the robot base linear velocity \( \mathbf{v}_{\text{base}} \in \mathbb{R}^3 \) from the feet in contact with the ground:

\begin{equation}
\mathbf{v}_{\text{base}} = -\frac{1}{n_c} \sum_{i=1}^{n} \mathbf{R}_b^w \left( \dot{\mathbf{p}}^i + \boldsymbol{\omega}_{\text{imu}} \times \mathbf{p}^i \right)
\label{eq:base_velocity_estimation}
\end{equation}
where \( \boldsymbol{\omega}_{\text{imu}} = [\omega_x, \omega_y, \omega_z]^\top \) is the IMU angular velocity vector, $n_{c}$ is the number of feet in contact with the ground,
and  \( \dot{\mathbf{p}}^i \) is the foot velocity in the world frame. 

The network for predicting the model error is a simple two-layer feedforward architecture with 256 hidden units per layer and ReLU activations. It is trained to minimize the mean squared error between the predicted correction \( \boldsymbol{\epsilon}_i^{\text{predicted}} \) and the true model error \( \boldsymbol{\epsilon}_i^{\text{true}} \), with respect to the loss function:
\begin{equation}
\mathcal{L}(\theta) = \frac{1}{N} \sum_{i=1}^{N} \left\| \boldsymbol{\epsilon}_i^{\text{predicted}} - \boldsymbol{\epsilon}_i^{\text{true}} \right\|^2
\end{equation}
Although the model error network is trained exclusively on real hardware data, its predicted corrections are incorporated during the RL policy training in simulation. Specifically, we use the covariance of the Kalman filter as written in Sec. \ref{sec:state_estimation} to inject realistic noise into the observations, encouraging the policy to learn robustness to real-world noise.

Finally, to improve smoothness, we apply a low-pass filter with smoothing coefficient $\lambda$ on the updated state from the Kalman filter:
\begin{equation}
\hat{\mathbf{x}}^{\text{filtered}}_t = \lambda \hat{\mathbf{x}}_t + (1-\lambda) \hat{\mathbf{x}}^{\text{filtered}}_{t-1}
\end{equation}

\subsection{Safety Guarantees using Reference Governor}
\label{sec:ref_gov}
\begin{figure}[t]
    \centering
\includegraphics[width=0.99\columnwidth]{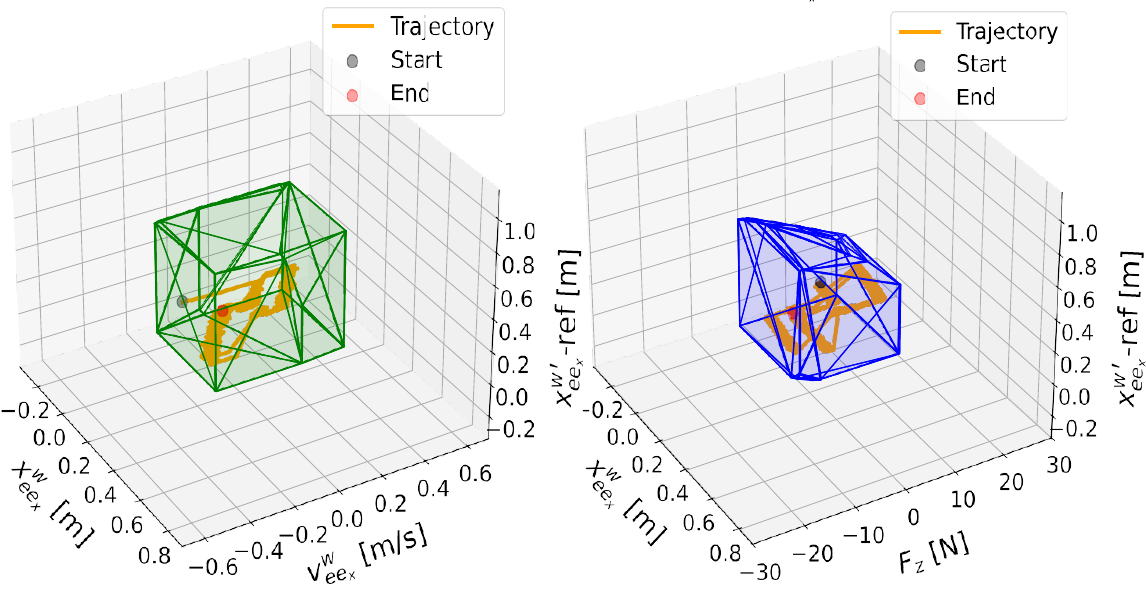}
    \caption{Maximal Output Admissible Set for the sample trajectory shown in Fig. \ref{fig:ref_governor_results}.} 
    \label{fig:moas_trajectories}
\end{figure}

To provide safety guarantees on Eq. \eqref{eq:admittance}, we employ a Reference Governor (RG), which ensures that the system operates within user-defined constraints by asymptotically adjusting the reference. As described in details in\cite{ref_gov1}, this method requires calculating the Maximal Output Admissible Set (MOAS), $\mathcal{O}_\infty$, which defines the entire set of initial states and references for which the system can operate without constraint violations:
\begin{equation}
    \mathcal{O}_\infty = \left\{ (x_0, v) \mid y(t) \in \mathcal{Y} \text{ for all } t \geq 0 \right\}
\end{equation}
where $x_0$ is the initial state of the system, $v$ is the reference input, $y(t)$ is the system output, and $\mathcal{Y}$ represents the set of permissible outputs with respect to constraints representing safety and performance requirements. In practice, the computation of the MOAS is an iterative process that verifies whether a given state and reference input remain within admissible bounds over a time horizon long enough to ensure convergence to a neighborhood of the equilibrium. The process begins by initializing position, velocity, wrench, and kinematic constraints, control gains, and discretization parameters. Grids are generated over position, velocity, reference position, wrench, and reference wrench to sample trajectories. For each grid combination, the algorithm simulates the system over discrete time steps using the control velocity $[\mathbf{v}_{ee}^{w}$, $\boldsymbol{\omega}_{ee}^{w}]$ to update position through Euler integration. If any constraint on position, velocity, wrench, or kinematics is violated during the trajectory, the grid point is discarded from the set. Otherwise, the corresponding values are added to $\mathcal{O}_\infty$. The final $\mathcal{O}_\infty$ comprises all initial conditions that can be controlled while satisfying all constraints. In our case, $\mathcal{O}_\infty$ is computed independently along the $x$, $y$, and $z$ axes for position, velocity, and force, and includes torque about the gripper base normal axis.

Once \( \mathcal{O}_\infty \) is obtained, it can be used to adjust the input reference set point at the current state to ensure a feasible control sequence exists that satisfies output constraints over an infinite horizon, i.e., $\mathbf{x}_{ee}^{'w}\text{, } \mathcal{W}^{'} = \mathcal{O}_\infty (o)$, where \( o = [\mathbf{x}_{ee}^{w}, \mathbf{x}_{ee\text{, ref}}^{'w},\mathbf{v}_{ee}^{w}, \mathcal{W}, \mathcal{W}_{\text{ ref}}^{'}] \) and $\mathbf{x}_{ee\text{, ref}}^{'w}, W_{\text{ref}}^{'}$ is the desired end-effector reference position and wrench. Note, if $o$ does not lead to constraint violation, according to  \( \mathcal{O}_\infty \), then $\mathbf{x}_{ee}^{'w}, \mathcal{W}^{'} = \mathbf{x}_{ee\text{, ref}}^{'w}, \mathcal{W}_{\text{ref}}^{'}$, \textit{i.e.}, references remain unchanged. To enable efficient real-time feasibility checking, we store discrete samples of \( \mathcal{O}_\infty \) in a \textit{KD-Tree}, allowing for fast nearest neighbor queries. Given a query state \( o_q \), we determine whether \( o_q \in \mathcal{O}_\infty \) or find the closest admissible state \( o^* \in \mathcal{O}_\infty \):
\begin{equation}
\label{kd_tree}
    o^* = \arg\min_{o \in \mathcal{O}_\infty} \|o - o_q \|
\end{equation}

If \( o \notin \mathcal{O}_\infty \) or Eq. \ref{kd_tree} is infeasible due to model error or significant disturbance, we hold \( \mathbf{x}_{ee}^{w}, \mathbf{v}_{ee}^{w}, \mathcal{W} \) constant and solve for the closest \( \mathbf{x}_{ee}^{*w} \) and \( \mathcal{W}^* \) that steer the system back into the admissible set, where now $\mathbf{x}_{ee}^{'w}, \mathcal{W}^{'}=\mathbf{x}_{ee}^{*w}, \mathcal{W}^*$.
\begin{figure*}[!t]
    \centering
    \includegraphics[width=6.8in]{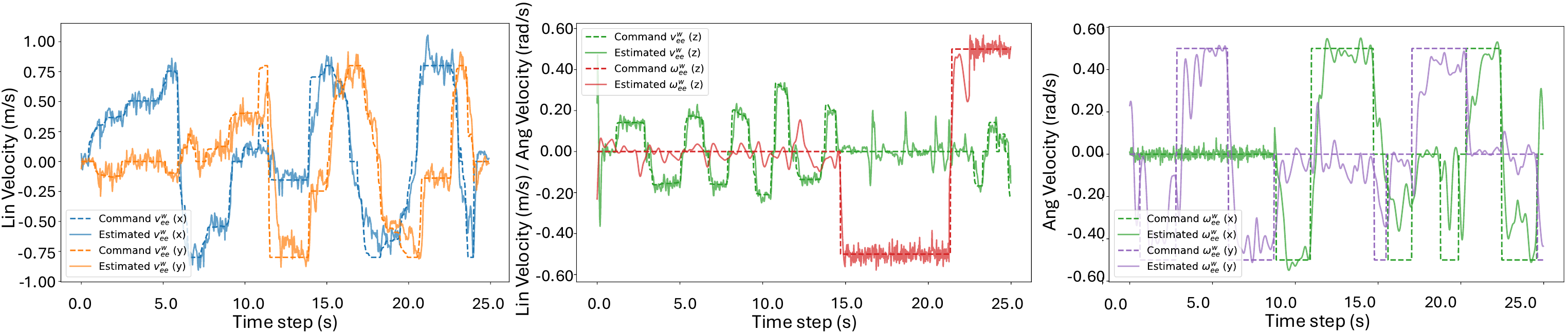}
    \caption{Velocity tracking results during human-robot collaboration task. Tracking results of the command end-effector linear velocity in $x/y$ directions is shown on the left, of linear and angular velocity in the $z$ direction in the middle, and of angular velocity in the $x/y$ directions on the right. The command linear and angular velocities are generated directly from the output admittance controller, which takes as input 6-DoF Force/Torque information. }  
    \label{fig:velocity_tracking}
\end{figure*}

\begin{figure}[t]
    \centering
    \includegraphics[width=0.99\columnwidth]{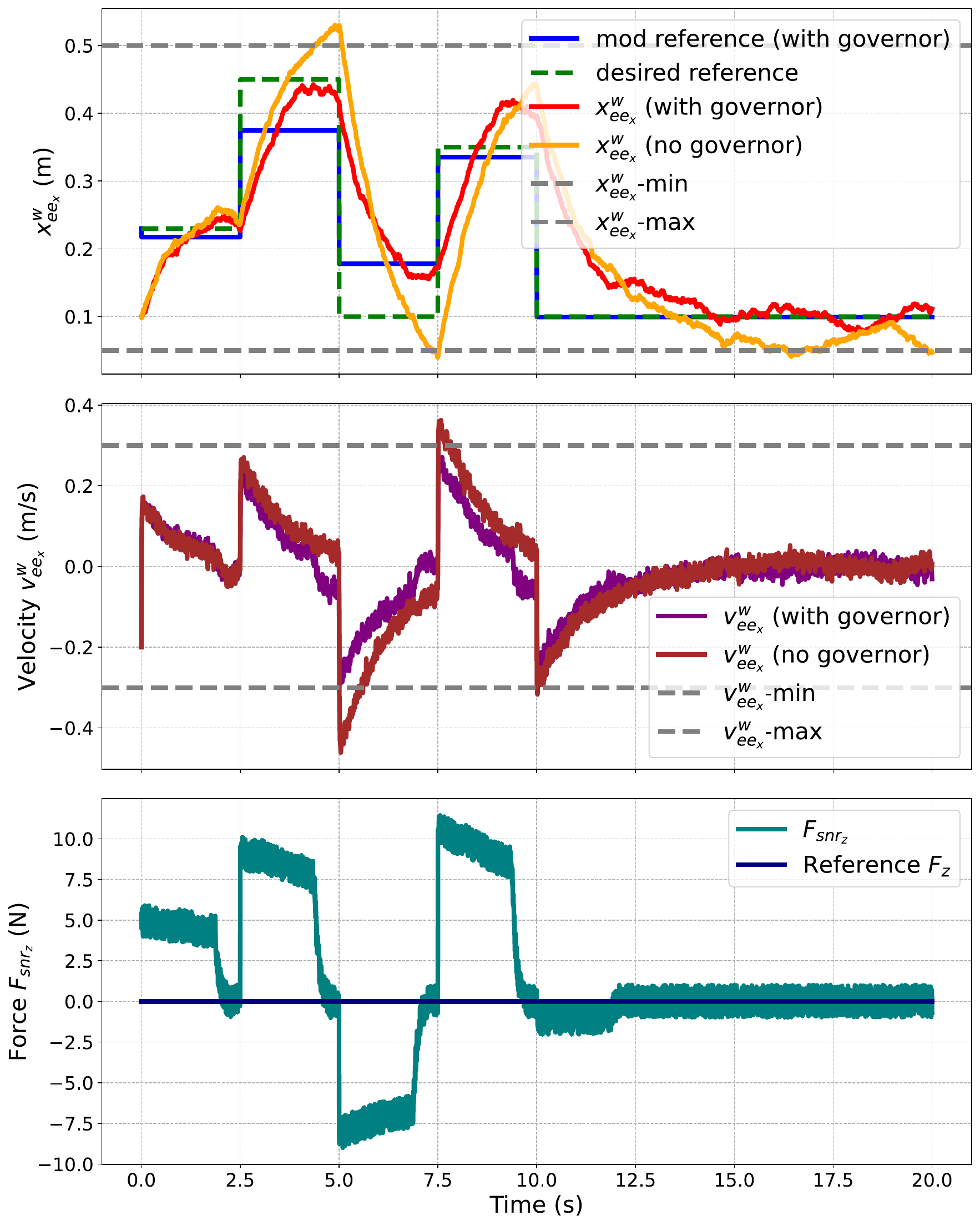}
    \caption{Example trajectory demonstrating the effect of the Reference Governor towards safety and constraint satisfaction.} 
    \label{fig:ref_governor_results}
\end{figure}

\begin{table}[ht]
\centering
\caption{Kalman Filter with Neural Network: \\Estimation Error Statistics}
\begin{tabular}{lccc}
\toprule
\textbf{Metric} & $\boldsymbol{\omega_x}$ (rad/s) & $\boldsymbol{\omega_y}$ (rad/s) & $\boldsymbol{\omega_z}$ (rad/s) \\
\midrule
MAE  & 0.0031 & 0.0013 & 0.0009 \\
RMSE & 0.0035 & 0.0017 & 0.0013 \\
STD  & 0.0018 & 0.0015 & 0.0013 \\
\midrule
\textbf{Metric} & $\boldsymbol{v_x}$ (m/s) & $\boldsymbol{v_y}$ (m/s) & $\boldsymbol{v_z}$ (m/s) \\
\midrule
MAE  & 0.0385 & 0.0235 & 0.0202 \\
RMSE & 0.0564 & 0.0297 & 0.0273 \\
STD  & 0.0511 & 0.0297 & 0.0272 \\
\bottomrule
\end{tabular}
\label{tab:ekf_stats_combined}
\end{table}

\section{Experimental Validation} 

\begin{figure}[t]
    \centering
\includegraphics[width=0.99\columnwidth]{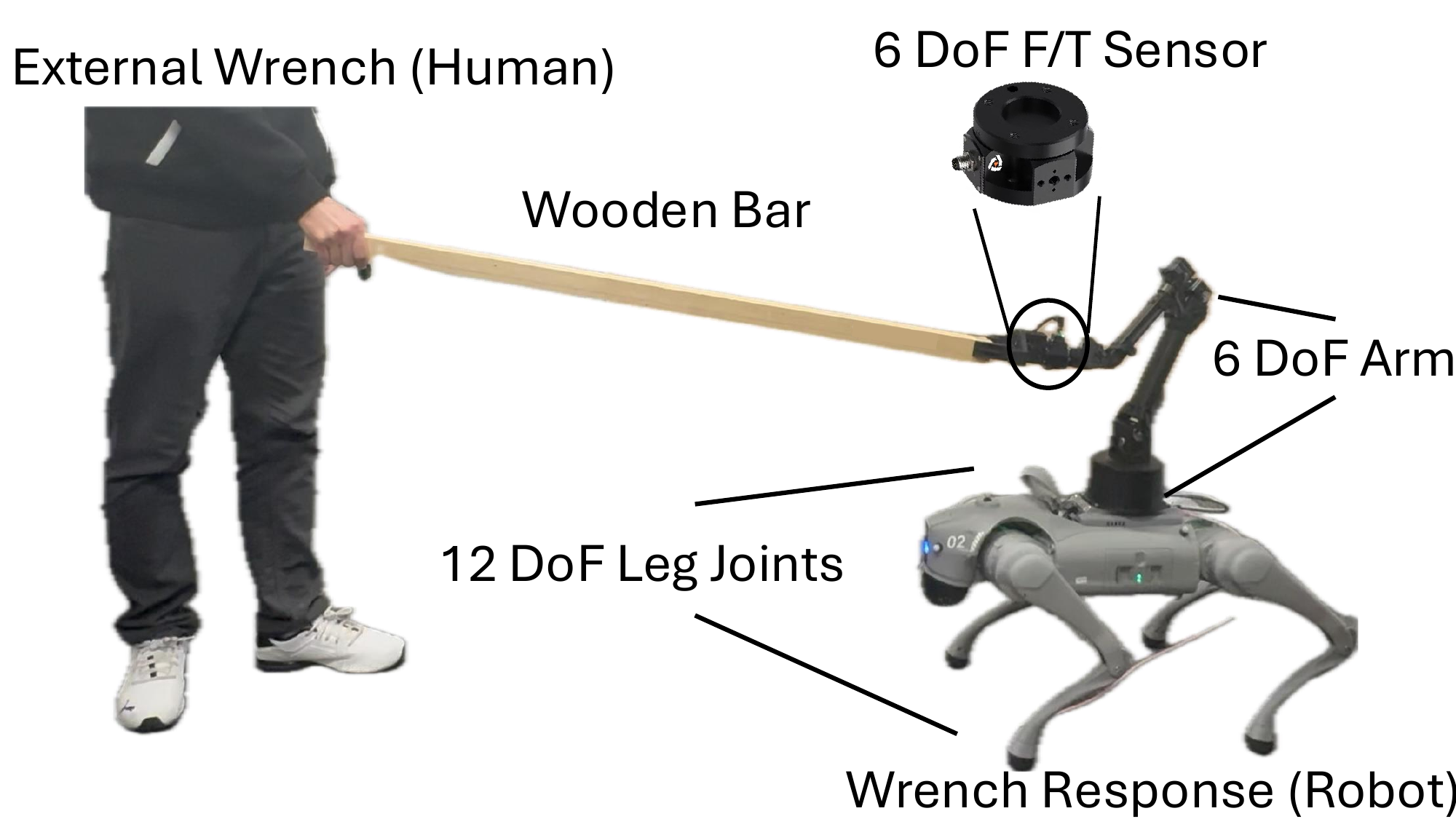}
    \caption{Hardware validation setup.} 
\label{fig:hardware_validation}
\end{figure}

We validate our framework on a Unitree Go2 quadruped robot equipped with a D1 6-DoF servo arm and a BOTA \cite{botasystems} F/T sensor attached at the arm's wrist (Fig. \ref{fig:hardware_validation}). The locomotion RL controller runs at 50 Hz, the arm’s admittance controller at 40 Hz (D1 arm motor send/receive frequency is limited at 40 Hz), and the state estimator at 500 Hz. Because the computation of the Reference Governor (RG) is offloaded offline, i.e., calculating the MOAS set, online deployment is fast as we simply query via the \textit{KD-tree} whether we are in the safe set or not, and modify the reference accordingly. We show the force controller’s ability to admit external wrenches in a human-robot collaboration task, where the robot and a human jointly held a wooden plank and moved it together in various directions, including over an obstacle, as shown in Fig. \ref{fig:intro}. The robot responded by generating the appropriate linear and angular velocity to follow the human’s motion based purely on the push/pull wrenches applied by the human on the end-effector. This type of human-robot collaboration task was achieved by setting the reference wrench in the RG input state to zero. Note, we do not impose a position reference in the $x$, $y$, and $z$ position of the end-effector, but do apply a constraint on the orientation reference, where the gripper is only allowed to rotate along the gripper base normal axis.

From over 25 seconds of loco-manipulation, Fig. \ref{fig:velocity_tracking} shows the end-effector's ability to track the commanded or interaction-driven linear velocity in $x$ and $y$ axes (left subfigure), linear and angular velocity in $z$ axis (middle subfigure), and angular velocity in $x$ and $y$ axes (right subfigure). The commanded velocity was computed via the force controller presented in Eq. \eqref{eq:compute_ee_vel_world}, based on the wrenches sensed by the F/T sensor at the wrist. Overall, we have a MSE~$\leq$~0.005 $\text{m}^2/\text{s}^2$ for linear and $\leq$~0.029 $\text{rad}^2/\text{s}^2$ for angular components, across all $x$, $y$, and $z$ axes. The results show accurate force tracking in all 6 DoF. Fig. \ref{fig:velocity_tracking} also shows a larger error in angular velocity between 15.0 s to 20.0 s for the $z$ angular velocity, with additional delay and noise in tracking the angular velocity in $x$ and $y$ axes. It should be noted that these errors may come from several sources, including actuator and sensor errors, external disturbances, and difficulty in tracking all 6 DoF at the same time. This can be directly affected by gain tuning of the admittance controller and having to prioritize some directional components over others.

We enforce stability/safety of our force controller through a RG, which modifies the desired reference (if needed) with a new reference to ensure constraint satisfaction and control stability. An illustration of the MOAS is given in Fig. \ref{fig:moas_trajectories} for position, velocity, reference position on the left, and for position, force, and reference position on the right. Using an example trajectory, as presented in more detail in Fig. \ref{fig:ref_governor_results}, we see that the start and end of the trajectory remain in the MOAS volume, indicating constraint satisfaction and recursive feasibility at all time steps. In Fig. \ref{fig:ref_governor_results}, we also compared following a desired trajectory, $x_{ee_x}^w$, and desired zero reference force, $F_{z}$ (to enable compliance), with and without a RG. For position tracking (top subfigure), without modifying the prescribed reference from 2.5 s to 5.0 s and from 5.0 s to 7.5 s, the controller can overshoot beyond the position constraint limits. This effect is apparently due to the quick change in the reference position at 5.0 s and 7.5 s, which causes a spike in the velocity seen at those time steps, violating the constraints (middle of subfigure). We show (bottom of subfigure), that our controller can successfully reduce the force back to zero, correcting for the force disturbance through the appropriate velocity commands. 

In Table \ref{tab:ekf_stats_combined}, we validate our state estimator from Sec. \ref{sec:state_estimation} for linear and angular velocity between the estimated state and ground truth from a motion capture system, using as metrics the Mean Absolute Error (MAE), Root Mean Squared Error (RMSE), and the Standard Deviation (STD).

\section{Conclusion}
Overall, we successfully demonstrated force-tracking loco-manipulation through a combined model and learning-based framework that admits user-applied wrenches and responds with the appropriate linear and angular velocities, achieving MSE values $\leq$~0.005 $\text{m}^2/\text{s}^2$ for the linear components and $\leq$~0.029 $\text{rad}^2/\text{s}^2$ for the angular components. The benefit of our model-based approach for the arm was shown by guaranteeing safety with the Reference Governor, which ensured that the output end-effector position and velocity always remained within the admissible set, abiding to position, velocity, and kinematic constraints. Finally, to compensate for errors in state estimation of the complex dynamical system of the legged arm robot, we employed neural networks within a Kalman filter---as improved estimation of velocity facilitates improved performance of our force controller. Future work will involve improving tracking for all DoF simultaneously, particular for torque, training our Neural Kalman filter on data where the robot is grasping various different objects to consider changes in weight distributions, and incorporating vision during the human-robot interaction task.

\balance
\bibliography{ref}
\bibliographystyle{ieeetr}

\end{document}